%% file: main_neurips.tex
\documentclass{article}

    \usepackage[preprint]{neurips_data_2023}

\usepackage[T1]{fontenc}    
\usepackage{hyperref}       
\usepackage{url}            
\usepackage{booktabs}       
\usepackage{amsfonts}       
\usepackage{nicefrac}       
\usepackage{microtype}      
\usepackage{xcolor}         
\usepackage{xspace}
\usepackage{enumitem}

\usepackage[skip=0.33\baselineskip]{subcaption} 
\usepackage{graphicx} 
\usepackage[capitalise]{cleveref} 
\usepackage{wrapfig} 
\usepackage[normalem]{ulem} 

\newcommand{\com}[1]{}
\newcommand{\resolved}[1]{}

\newcommand{\newparagraph}[1]{\noindent {\bf #1}}
\renewcommand{\newparagraph}[1]{\paragraph{#1}}
\newcommand{\merlot}{MERLOT Reserve\xspace}

\title{Read, Look or Listen? What's Needed\\ for Solving a Multimodal Dataset
}

\author{%
\textbf{Netta Madvil}
\qquad\textbf{Yonatan Bitton}
\qquad\textbf{Roy Schwartz}
   \\
  School of Computer Science,   The Hebrew University, 
  Jerusalem, Israel\\  \texttt{\{netta.madvil,yonatan.bitton,roy.schwartz1\}@mail.huji.ac.il} 
}

\begin{document}

\maketitle

\begin{abstract}
\input{sections/0-abstract}

\end{abstract}

\section{Introduction}
\input{sections/1-introduction}

\section{Mapping Instances in Multimodal Datasets} \label{ssec:three}
\input{sections/2-Humans}



\section{A Case-study: TVQA}
\label{sec:tvqa}
\input{sections/3-TVQA_Analysis}


\section{A New Multimodal Test Set for TVQA. }
\input{sections/4-new_test}

\section{Related Work}
\input{sections/5-related_work}

\section{Limitations}
\input{sections/6-limitations}

\section{Conclusion}
\label{sec:limitations}
\input{sections/7-conclusion}

\section{Acknowledgements}
\input{sections/8-acknowledgements}

\bibliographystyle{unsrtnat}
\bibliography{references}


\newpage
\appendix

\section{Appendix}
\input{sections/Appendix}

\end{document}

%% file: sections/0-abstract.tex
The prevalence of large-scale multimodal datasets presents unique challenges in assessing dataset quality. 
We propose a two-step method to analyze multimodal datasets, which leverages a small seed of human annotation to map each multimodal instance to the modalities required to process it. Our method sheds light on the importance of different modalities in datasets, as well as the relationship between them.
We apply our approach to TVQA, a video question-answering dataset, and discover that most questions can be answered using a single modality, without a substantial bias towards any specific modality. Moreover, we find that more than 70\% of the questions are solvable using several different single-modality strategies, e.g., by either looking at the video \textit{or} listening to the audio, highlighting the limited integration of multiple modalities in TVQA.
We leverage our annotation and analyze the \merlot model, finding that it struggles with image-based questions compared to text and audio, but also with auditory speaker identification. Based on our observations, we introduce a new test set that necessitates multiple modalities, observing a dramatic drop in model performance. Our methodology provides valuable insights into multimodal datasets and highlights the need for the development of more robust models.

%% file: sections/1-introduction.tex
AI models are highly affected by their training data. As a result, understanding what's inside these datasets is important, both in order to improve the underlying models, and to mitigate their biases~\citep{dodge-etal-2021-documenting}. Nonetheless, the scale of modern datasets makes such an analysis challenging. To tackle this task, previous work has primarily focused on understanding dataset characteristics, such as their outliers~\citep{Carlini2019DistributionDT}, the learnability of different instances~\citep{Swayamdipta2020DatasetCM,Nam2022SpreadSA,Siddiqui2022MetadataAU},  and the biases~\citep{Luccioni2023StableBA} and mislabels~\citep{Talukdar2021TrainingDB} they contain. 


In this work we analyze a specific family of AI datasets---multimodal datasets~\citep{Tapaswi2015MovieQAUS, Ye2017VideoQA,lei2018tvqa},
which contain information from different modalities, such as text, images, and audio. 
An important question regarding these datasets is the relative importance of each modality and how it manifests within dataset instances.

\begin{figure}[t]
\centering
\begin{subfigure}{0.45\linewidth}
\includegraphics[width=\linewidth,scale=0.5, trim={10cm 4cm 10cm 2cm},clip]{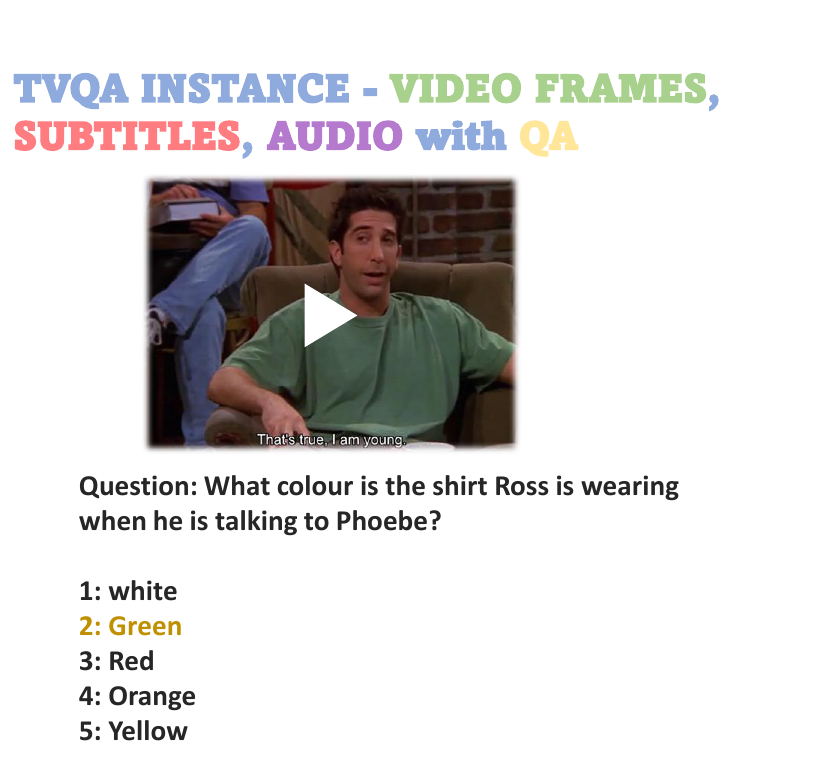}
\caption{TVQA instance (correct answer marked).\label{fig:tvqa_example}}
\end{subfigure}%
\quad
\begin{subfigure}{0.45\linewidth}%
\includegraphics[width=\linewidth,scale=0.5, trim={10cm 4cm 10cm 2cm},clip]{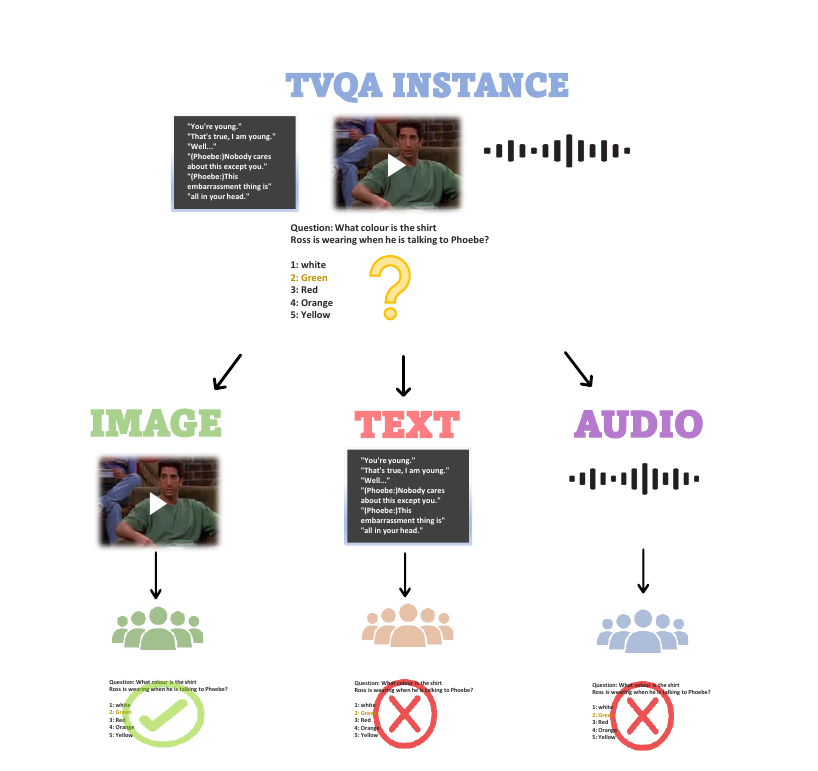}
\caption{Our annotation framework.}
\end{subfigure}
\caption{\label{fig:first} 
Our annotation framework for identifying the modalities required for processing multimodal instances.
(a) a TVQA instance (b) our framework: we break each instance into three groups: audio, text, and image. Separate groups of crowdworkers try to answer the questions in each group. 
Here the workers are able to solve the question using image alone, but not using text or audio, indicating that it only requires the image.}
\end{figure}

We present a two-step method, which maps each instance in multimodal datasets to the subset of modalities required to process it. Our method relies on a small seed annotation step, which is later expanded to the full dataset using classification tools.  For example, \cref{fig:tvqa_example} shows an instance from TVQA~\citep{lei2018tvqa}, a video question-answering dataset, which contains a question regarding the clothes worn by one of the characters. This question could be solved by viewing the image, even without the audio or subtitles. In contrast, hiding the video and showing the other two modalities makes it impossible to solve. 
Our method allows analyzing multimodal datasets, while gaining insights into the underlying relationships and dependencies between the different modalities. It also allows  assessing model capabilities on the different modalities. 

We apply our approach to TVQA, 
observing a few interesting findings. First, we validate previous findings~\citep{winterbottom2020modality}, and show that 99\% of the questions can be answered using a single modality. However, unlike that fully-automated work, which found a bias in the data towards the text modality, our method, which relies on human annotation, shows no substantial bias towards any modality. Second, we find that more than 70\% of the questions are solvable by two or more modalities separately, and more than 15\% using each of the three modalities
.
We then leverage our analysis to study a given model's performance on the TVQA dataset, by running the \merlot model~\citep{Zellers2022MERLOTRN} on different instances requiring different modalities. We find that this model generally struggles with image-based questions, but also with  questions requiring audio speaker recognition. 

Finally, based on our observations, we collect a new test set of 150 questions that cannot be answered using any single modality. We find that \merlot performs dramatically worse on these questions compared to the original validation set (41\% vs.~83\% accuracy), suggesting that it struggles with questions that require multiple modalities. We hope these findings will inspire others to develop methods for training more robust multimodal models.

\setlist[itemize]{itemsep=0pt, topsep=0pt}

\noindent {\bf Contributions.} In summary, our main contributions are as follows:
\begin{itemize}[leftmargin=12pt]
    \item A novel two-step method for mapping multimodal instances to the required modalities.
    \item An analysis of modality importance and manifestation within instances of the extended TVQA dataset, providing insights into the characteristics of the dataset.
    \item Assessment of \merlot capabilities and biases on the TVQA dataset, revealing the model's performance with different modalities.
    \item A new challenging test containing questions that require multiple modalities.
\end{itemize}

%% file: sections/2-Humans.tex
Our goal is to map the instances in a given multimodal dataset to the modalities required for processing them. To accomplish this, we present a simple two-step annotation methodology. We first sample a subset of the dataset, and use human annotators to map each instance in it to the subset of modalities required to process it. We then train several classifiers, one per modality sub-group, on the collected annotations. We apply the resulting classifiers to the full dataset, resulting in a mapping of each instance to the a subset of the modalities it requires. We turn to describe both parts.

\newparagraph{Collecting small seed annotations.}
We start by sampling a subset of the data, to be used for our seed annotation. We present human annotators with different subsets of modalities for each instance, recording their responses. We first collect annotations of a single modality and then gradually increase their number (e.g., both image and audio). This allows us to identify which subsets of modalities are sufficient for processing each instance.

More formally, consider a dataset $D$ containing a set of modalities $M$ of size $|M|$. We sample a subset $D' \subset D$. For an instance $i \in D'$ and modality $m \in M$, we mark the version of $i$ that only contains information from modality $m$ as $i_m$.\footnote{E.g., in a movie dataset, $i_m$ could be the sound component of a given movie.} We then ask human annotators to label $i_m$ for each instance $i\in D'$, and modality $m \in M$. Naturally, some instances $i_m$ cannot be solved without access to some modalities. As a result, annotators are likely to perform at chance level in these cases, indicating that instance $i$ is insolvable with modality $m$ alone. In order not to contaminate the process, we divide the group of annotators between the different modalities, such that no annotator sees the same instance more than once with different modalities.
After considering single modalities, if a sufficient portion of the instances cannot be solved using any single modality, we continue annotating them with pairs of modalities, and if necessary triples, quadruplets, etc.
The resulting annotations allow us to analyze and characterize the underlying dataset, by asking questions such as which instances can be solved using individual modalities; which require more than a single modality to process; which can be processed by more than one modality; etc.
Aggregating these annotations allows us to map the different regions of a given dataset, and visualize it.

As an example, consider \citet{Zellers2022MERLOTRN}'s version of TVQA~\citep{lei2018tvqa}, which contains three modalities---audio, text, and image. For each modality $m$, participants are shown the $m$ part of a scene (without access to the other modalities), and a multi-choice question. \cref{fig:first} shows Ross from the TV show ``Friends'' wearing a green shirt while talking to Phoebe. The corresponding question is ``What colour is the shirt
Ross is wearing when he is talking to Phoebe?'', which can be answered using the visual signal, but not without it.

\begin{figure}[t]
\includegraphics[width=\linewidth, scale=1, trim={5cm 7.5cm 4cm 6cm},clip]{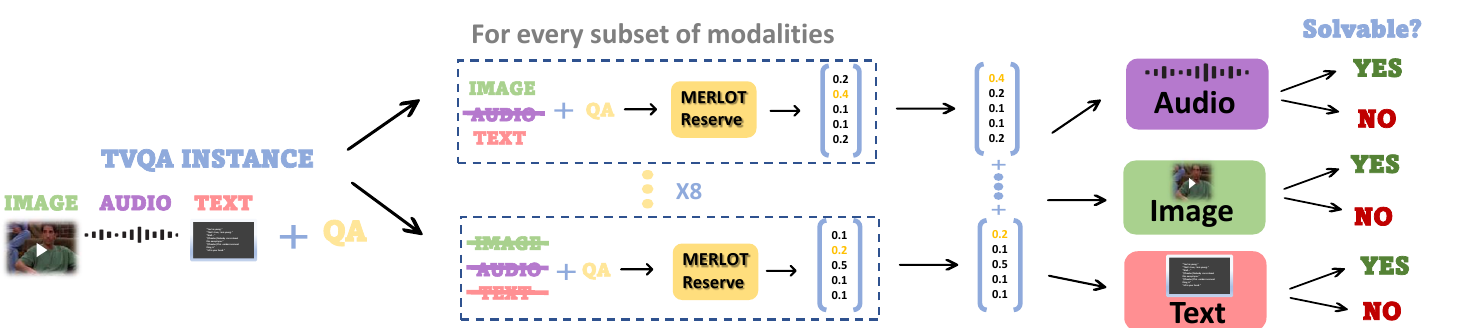}
\caption{\label{fig:third} Illustration of our annotation of the full dataset, exemplified TVQA. An instance is mapped into eight ($=2^{|M|}$) different combinations of modalities. 
These subsets are fed to a model trained on the original TVQA training set, which computes the softmax output probabilities of possible answers. These probability vectors, sorted such that the gold label's probability is first`, are then concatenated to form the input vector for each classifier, which is trained to predict the answerability of a given instance using a specific modality.}
\end{figure}

\newparagraph{Expanding to the full dataset.} We next extend our human annotations to the entire dataset. To do so, we train $|M|$ classifiers, one for each modality $m$, on the annotations collected above, i.e.,  predicting whether or not a given instance is solvable using $m$ only. 
To generate the features for our classifiers, we start by fine-tuning a model on the original training set of the given dataset (including all modalities). Next, we select all subset combinations of modalities ($2^{|M|}$ combinations), and for each one generate a version of the validation set with the given subset masked out. We then apply the trained model on the masked instances and extract the softmax layers' outputs obtained from each modality subset. 
We concatenate these output vectors to create an input vector of size $2^{|M|}*|L|$, where $L$ is the label space
. We then train a random forest~\citep{ho1995random} classifier for each modality, using these input features, and the annotated labels collected above. We apply the trained classifiers to the full validation set, and obtain a silver annotation.\footnote{A similar process can be applied to annotate the training set, for details see \cref{app:halves_train}.} See \cref{fig:third} for illustration.

%% file: sections/3-TVQA_Analysis.tex
We apply our method to the TVQA video question answering dataset~\citep{lei2018tvqa}.
Below we describe our data collection (\cref{sec:dataset_experimental_setup}); how our results provide insights on both the TVQA dataset (\cref{sec:dataset_analysis}); and a model trained on this dataset (\cref{sec:dataset_model_analysis}).

\subsection{Data Collection}\label{sec:dataset_experimental_setup}

\newparagraph{TVQA}
 contains 150K question-answer pairs over 6.5K video clips from six popular TV shows. 
The questions in the dataset cover a broad range of topics, including object recognition, scene understanding, and story comprehension. The dataset also includes a set of multiple-choice questions, where each question has five possible answer choices. The dataset was originally introduced as a two-modalities dataset (video frames and subtitles), but we consider \citet{Zellers2022MERLOTRN}'s version, which includes a third modality, namely speech. 

\newparagraph{Humans annotations.}
We hire native English-speaking workers from Amazon Mechanical Turk to answer questions based on a specific modality of the input. We develop qualification tests to select high-quality annotators and divide them into three separate groups, one per modality. Following \citet{chen-etal-2020-hybridqa} and \citet{Castro2020LifeQAAR}, we also require participants determine whether they think the question can be answered based solely on specific modality input. However, we find this approach to be less reliable---35\% of the questions marked as unsolvable were, in fact, solvable. As a result, we rely on the annotators accuracy in order to determine whether a question is answerable using a subgroup of modalities, but use the yes/no information for monitoring the quality of their annotations.\footnote{Workers who consistently marked `yes' or `no' were excluded, as well as those who showed low agreement with the rest of the group.} 
To prevent annotators from answering based on memory of already seeing that particular scene, we add a checkbox for them to indicate whether they had seen the scene before, and omit those answers. 

We select a set of 650 examples from the validation set of TVQA, all from the TV show ``Friends'', containing 500 randomly selected examples, and  150 additional instances that are most ``sensitive'' to the model for each modality (See \cref{app:non-random}). We break them down to 75\%/15\%/10\% for training/validation/test, respectively. We additionally collect 150 examples from ``House M.D'', to serve as an out-of-distribution (OOD) test set.   We show each annotator a single modality, and use five annotators for each (instance,modality) combination. The final label is determined by majority vote. See \cref{app:workers-analysis} for more information about the annotation metrics and further analysis.

\newparagraph{Expansion to the full dataset.}
For each of the TVQA modalities, we train a different classifier, each based on a frozen finetuned \merlot model~\citep{Zellers2022MERLOTRN}, which achieves a relatively high level of accuracy (83\%) on the original TVQA dataset. 
The classifiers are trained on our human annotated labels,  with balanced class weights for each modality. An illustration of our three classifiers as applied to TVQA is presented in~\cref{fig:third}. 
We conduct a hyperparameter search over the basic parameters of Random Forest such as the number of trees in the forest and the maximum depth of each tree. We report test results on the top-performing model on our validation set. See \cref{app:scorer-analysis} for more details.

\newparagraph{Classification results.} 
\cref{tab:three_tasks_results} shows our results. Test results range between 74-81\%, on average 12\% higher than a majority baseline. Interestingly, our OOD test results are on par, and sometimes higher than the ID test results, indicating that our annotations generalize to other domains. We also test the effect of the training set size, by training the classifiers with 30\%, 50\% and 70\% of our training data. Our results (\cref{tab:three_tasks_results_different_training_size} in~\cref{app:data-size-analysis}) show that training on as few as 30\% of our original size yields performance comparable to our full training set. This finding not only streamlines the data collection process, but also suggests that strong results can be achieved without further annotation efforts.

\begin{table*}[t]
\caption{\label{tab:three_tasks_results}
Validation, test, and OOD accuracy of our classifiers for predicting the solvability of  TVQA instances based on a single modality, compared to a majority baseline.
}
\centering
\begin{tabular}{@{}p{0.001cm}p{1.8cm}cccccccccc@{}} \toprule
\multicolumn{2}{l}{} & 
\multicolumn{3}{c}{Image} & 
\multicolumn{3}{c}{Text} & 
\multicolumn{3}{c}{Audio}  
\\ \cmidrule(l){3-5} \cmidrule(l){6-8} \cmidrule(l){9-11} 
\multicolumn{1}{l}{}     & Approach & Val  & Test  & OOD   & Val  & Test  & OOD  & Val  & Test  & OOD \\ \midrule
& Majority             
& \phantom{0} 72              & \phantom{0}72                & \phantom{0} 61              & \phantom{0} 61              & \phantom{0} 61               & \phantom{0} 69               
& \phantom{0} 69              & \phantom{0} 69               & \phantom{0} 71  \\


& Classifier
& \phantom{0}89              & \phantom{0}81                &
\phantom{0}80              & 
\phantom{0} 82              & 
\phantom{0} 74               & 
\phantom{0} 80               & 
\phantom{0} 81              &
\phantom{0} 76               & \phantom{0}77  \\
                            
\bottomrule
\end{tabular}
\end{table*}

\subsection{Dataset Analysis}\label{sec:dataset_analysis}
We use our classifiers to analyze the validation set of the TVQA dataset. A similar analysis of the training data appears in \cref{app:halves_train}, and shows similar results. An analysis based on our seed annotation also yields similar results, see~\cref{app:tvqa_analysis_based_on_workers}.

\begin{figure}[htbp]
\centering
\begin{subfigure}[t]{0.48\textwidth}
\includegraphics[width=\textwidth,scale=2, trim={1cm 0.3cm 1.1cm 0.5cm},clip]{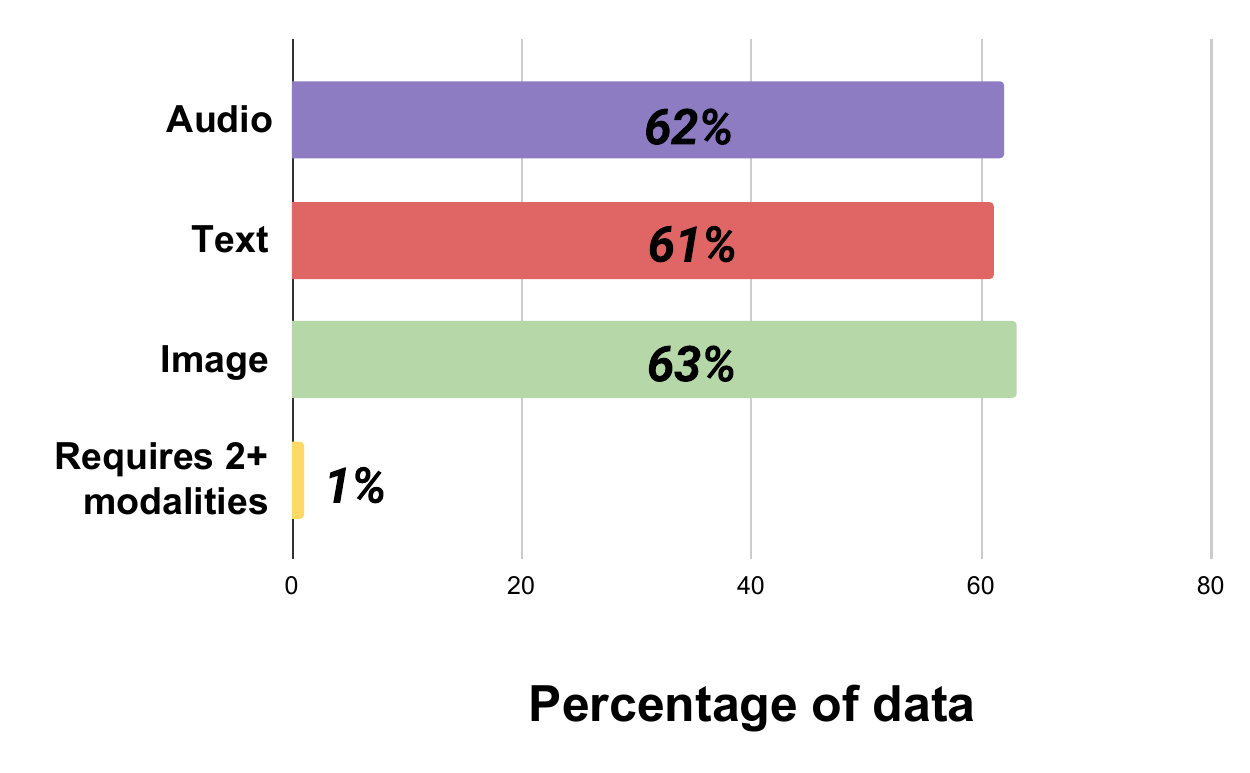}
    \caption{\label{Histogram val} Proportions of questions answerable using a single modality (top three bars), as well as that of questions unanswerable using any single modality (final bar).}
\end{subfigure}
\hfill
\begin{subfigure}[t]{0.48\textwidth}
    \includegraphics[width=\textwidth,scale=6, trim={4cm 2.5cm 7cm 1cm},clip]{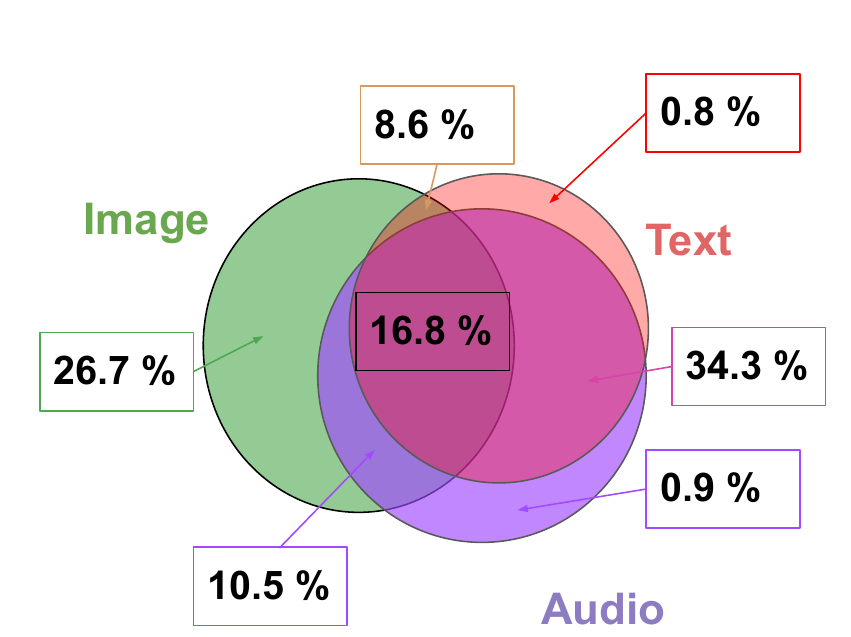}
        \caption{\label{venn val}
Each circle in the graph represents a portion of the data answerable using a particular modality. Overlapping areas mark the proportions of instances solvable by either modality separately.}
\end{subfigure}
\caption{An analysis of the validation set of the TVQA dataset.}
\label{fig:validation-splits}
\end{figure}

\newparagraph{99\% of the questions are solvable using a single modality.}
\cref{Histogram val} shows the proportions of questions answerable with each modality, as well as the fraction of questions unanswerable using any single modality. Our results confirm previous results~\citep{winterbottom2020modality} and show that almost 99\% of the questions could be solved correctly using a single modality. As a result, we do not further annotate groups of more than one modality.

\newparagraph{Questions are balanced across modalities.}
We observe that similar proportions of the questions could be solved by each of the different modalities---63\%/61\%/62\% using image/text/audio, respectively. Prior work~\citep{winterbottom2020modality} used partial-input models and identified a bias towards the text modality in this dataset. Our method, relying on a seed annotation, leads to a different conclusion and shows no evidence of such bias.

\newparagraph{Many questions could be solved by more than one modality.}
The values in~\cref{Histogram val} indicate a large overlap between the instances answerable by the different modalities (as the sum of all bars largely exceeds 100\%).
To study this overlap, we plot (\cref{venn val}) a Venn diagram, where each circle represents the partitions of instances answerable with a different modality, and overlapped areas represent instances solvable by either modality separately. 
Our results show that more than 70\% of the instances are solvable using two or more modalities, and more than 15\% using each of the three modalities, further indicating that many TVQA instances do not require integration of different modalities.

\newparagraph{Large overlap between audio and text.} We observe a high overlap between questions answerable using only text and those answerable using audio alone. 
This indicates that most questions focus only on what is said in the scene (typically reflected in both the audio and the subtitles), rather than additional information, e.g., non-verbal sounds such as beeps, horns, or music. It also indicates few to no questions with blurry or noisy speaking that is hard to understand without the subtitles. 
In contrast, 26.7\% of the questions are answerable  using the image, but not the other modalities.

\subsection{Analysis of Dataset/Model Interaction} \label{sec:dataset_model_analysis}
\begin{wraptable}{R}{0.52\textwidth}
\vspace{-12pt}
\caption{\label{tab:scorer rep}Accuracy of various versions of the \merlot model, each time masking different modalities (columns), and evaluated on different data splits of TVQA (rows). Columns: that modalities shown: (\textbf{I})mage, (\textbf{A})udio, (\textbf{T})ext, \textbf{A+T}: audio and text. \textbf{base}:baseline model. 
The last block shows model performance on `who' questions.
}
\centering
\begin{tabular}{@{}cccccc@{}} \toprule
\textbf{Data\textbackslash  Input}  & \textbf{base} & \textbf{I} &  \textbf{A} &  \textbf{T}
&\textbf{A+T} 
\\ \midrule
All  & 81  &  57& 68 & 69 & --\\
Answerable-all  & 88  & 62& 78 & 85 & --\\
\midrule
Only Image & 74  & 75 & -- & --& --\\ 
Only audio--text & 94  & --  & 86  & 87 & 96\\
\midrule 
`who' questions & 88  & 71 & 45 & 95 & --\\ 
\bottomrule
\end{tabular}
 \vspace{-12pt}
\end{wraptable}

We have so far presented a comprehensive analysis of the TVQA dataset, marking for each instance the modalities required for processing it. We now set to analyze a model trained on this dataset, in order to identify the modalities on which the model struggles with, and gain insights into the interplay between modalities when solving the task.
We focus on \merlot~\citep{Zellers2022MERLOTRN}, the only publicly available model finetuned on TVQA  with all three modalities (to the best of our knowledge).\footnote{\url{https://github.com/rowanz/merlot_reserve}}
We apply the model to several data splits based on our  annotations, and describe our experiments and takeaways below.

\newparagraph{\merlot reads better than it sees.} We first consider the finetuned \merlot model, and perform inference with it on three versions of the validation set (\textbf{All} row in \cref{tab:scorer rep}), each time masking 2/3 modalities, and leaving only one source of input (\textbf{I}mage, \textbf{A}udio and \textbf{T}ext columns). We compare these performance scores to the model's baseline performance (\textbf{base} column).
Our results show that, unsurprisingly, model performance drops in all cases. However, this decrease is not uniform across modalities: using only the image modality leads to the largest drop (24\%), compared to 12--13\% for audio and text.

To further investigate this trend, we repeat this experiment with the questions answerable by each of the modalities (i.e., the center of \cref{venn val}), which are considered somewhat easier, as they can be solved using multiple cues from the different modalities. Our results (\cref{tab:scorer rep}, \textbf{Answerable-all}) show a similar trend for image and audio, but the text-only version is only 3\% behind the baseline model. 
Our findings indicate that the model faces difficulty in using the image modality to answer questions. This aligns with previous work, which highlighted the underutilization of the image modality in multimodal tasks~\citep{Zhang2015YinAY, Goyal2016MakingTV,Jabri2016RevisitingVQ,  Hassantabar2018VisualQA, Bitton2021DataEM, Dancette2021BeyondQB}.

\newparagraph{\merlot struggles with image-based questions.}
We have so far shown that \merlot struggles when given access only to the visual component. We next return to the full model (i.e., no masking, \textbf{base} column in \cref{tab:scorer rep}), and examine whether this trend translates to difficulties in processing questions that require visual information. We consider questions labeled to be answerable by our classifiers using the image modality only, compared to those answerable only by either text or audio.\footnote{Due to the high overlap between audio and text observed in \cref{sec:dataset_analysis}, we compare image-only questions to audio-text-only questions.} Our results (middle block of \cref{tab:scorer rep}) show a substantial gap (20\%) in favor of the latter, indicating that the model struggles with questions that require the image modality, while it succeeds on those that are answerable by audio or text.

\newparagraph{A training dynamics analysis.}
Results observed so far indicate that questions that require processing visual information are harder for \merlot compared to those requiring text or audio. To further validate this hypothesis, we compute the training dynamics~\citep{Swayamdipta2020DatasetCM} of the TVQA dataset using \merlot. This is an alternative method for mapping a dataset into different regions: \textit{easy-to-learn}, \textit{hard-to-learn} and \textit{ambiguous}. We fine-tune the \merlot model with the same parameters used in the original paper, training it for three epochs while calculating the mean and variance of the softmax output probability of the gold label for each instance.
We then select the top 50\% of our annotated instances with the highest variance (i.e., 50\% most \textit{ambiguous} instances in \citet{Swayamdipta2020DatasetCM}'s terminology). \cref{tab:traning_dynamics} presents the answerability proportions for each modality in the 50\% most ambiguous questions, and compares them to the answerability proportions of all annotated data (from \cref{Histogram}). Interestingly, we notice a rise in the proportion of image-based questions in the 50\% most ambiguous questions, whereas the proportion of audio/text-based questions decreases. 
As the 50\% most ambiguous questions are considered more challenging~\citep{Swayamdipta2020DatasetCM}, these results support our previous findings---\merlot faces more difficulty in answering questions that require visual information compared to other modalities.

\begin{wraptable}{r}{0.44\textwidth}
\vspace{-12pt}
\caption{\label{tab:traning_dynamics} Training dynamics analysis of the validation set of TVQA. For each data portion (rows), we calculate the proportion of questions answerable by each modality on it. \textit{all}: all annotated data, \textit{most ambig.}: 50\% most ambiguous examples. 
The prevalence of image-based ambiguous questions indicates the model's difficulty with the image modality.
}
\centering
\begin{tabular}{@{}ccccccc@{}} \toprule
\textbf{Data}   &  \textbf{A} (\%)  &  \textbf{T} (\%)  & \textbf{I}  (\%) 
\\ \midrule 
all & 63 & 61 & 63 \\ 
most ambig.
& 48 & 43 & 83\\
\bottomrule
\end{tabular}
\end{wraptable}

\newparagraph{\merlot's limitations in speaker recognition.}
We turn to further analyze the \merlot model fine-tuned on TVQA, by evaluating its performance on a particular type of question: ``who'' questions, where the correct answer is one of the main characters in the TV show. E.g., the question ``Who is Monica talking to when she is upset and crying?''. To answer such questions, all three modalities---image, audio, and text---can be used, as the characters can be recognized through their looks, their voices, or their names as they appear in the subtitles. We therefore check whether the model is equally capable of using these modalities. We run the fine-tuned \merlot model, and use masking to create image-only, audio-only, and text-only versions of the model to these questions, as described above. As some ``who'' questions might be answerable by only one modality (e.g., if the target character does not speak during the scene), we only consider the proportion of ``who'' questions that are answerable using each modality according to our annotation. Our results (\cref{tab:scorer rep}, last block), indicate that the audio modality has a substantially lower score compared to text and image when answering these questions. This indicates that the model struggles in speaker recognition. It is worth noting that in the text modality, the answer to ``who'' questions does not require any memory or learning of the characters since the character's name is usually explicitly written. In contrast, the image modality is similar to audio in this sense, where the model needs to recognize the main characters visually, based on its training. The gap between the audio and image modalities emphasizes the model's challenges in recognizing the main characters through speech, compared to visually.

%% file: sections/4-new_test.tex



\begin{wrapfigure}{r}{6.75cm}
\vspace{-20pt}
\includegraphics[scale=0.3, trim={7cm 0cm 1cm 0.5cm},clip]{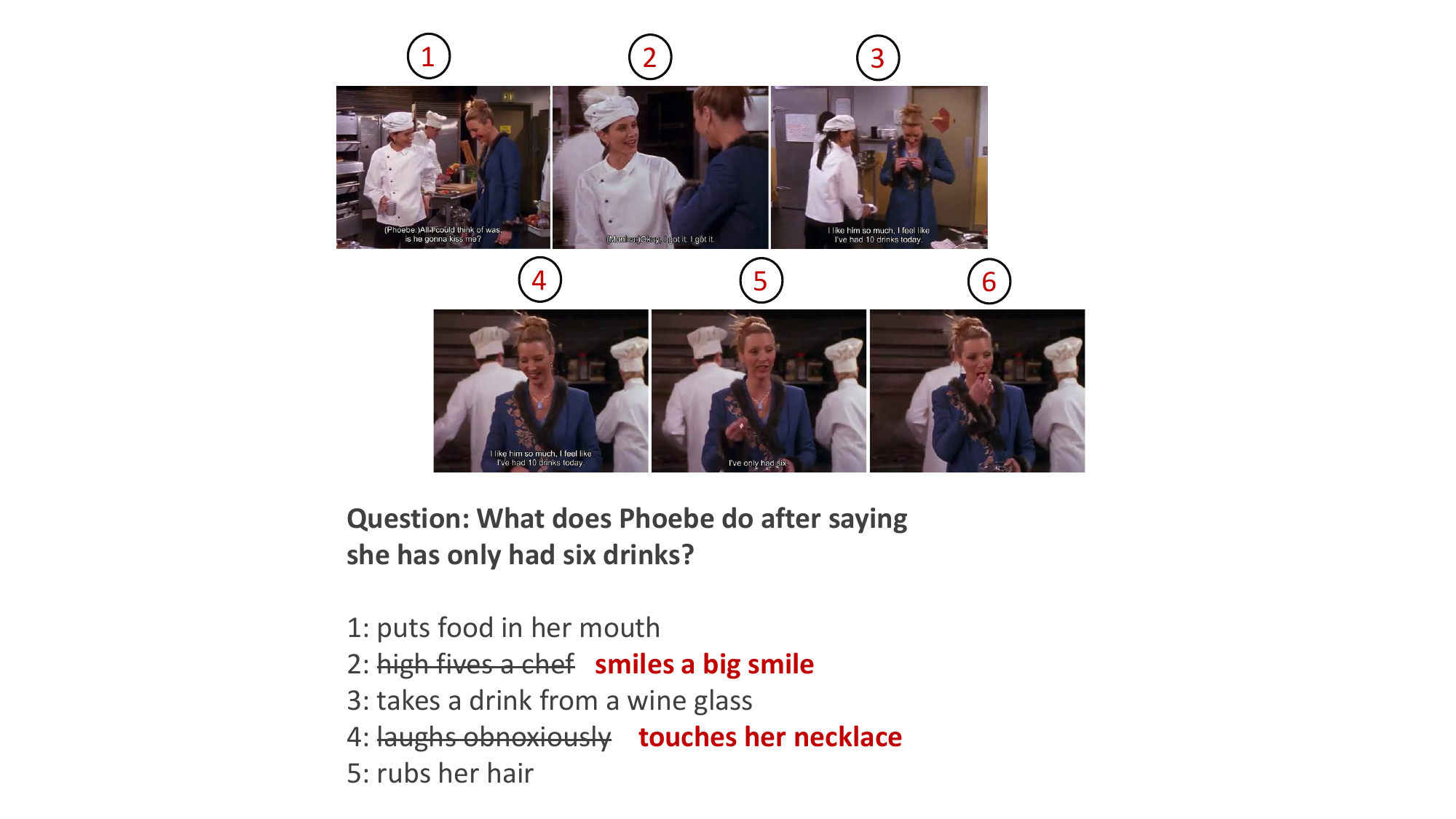}
\caption{\label{fig:mmqa} An original TVQA question that relies on visual cues alone, along with improved distractors that now require both image and sound/text. The video frames appear  in chronological order from left to right.}
\vspace{-20pt}
\end{wrapfigure}

Our analysis in~\cref{sec:dataset_analysis} has shown that TVQA contains almost no questions that require more than one modality. To test the impact of this deficiency on models trained on TVQA, we crowdsource a set of 150 questions that require multiple modalities (see \cref{app:mmqa} for information about the data collection). Another group of workers then filter out questions that are either not multimodal or insolvable. We observe that one approach used by our workers is to modify distractors in existing TVQA questions, in order to force the model to use multiple modalities. 

For example, in \cref{fig:mmqa}, we have the question ``What does Phoebe do after saying she has only had six drinks?''. The original correct answer (``puts food in her mounth''), is the only option describing an action performed in the video, and as a result an image-only model is able to answer it correctly. Our annotators modify this question by adding distractors such as ``smiles a big smile'', an action Phoebe performs while saying something else. See \cref{app:mmqa} for more examples.

We evaluate the pre-trained \merlot model on the collected questions, observing that it performs substantially worse on them compared to its performance on the original validation set---41\% vs.~83\%. This indicates that the model struggles with questions that require more than one modality. In order to make models more robust, future work will involve collecting a larger set of such questions, both for training and evaluation.

%% file: sections/5-related_work.tex
\newparagraph{Dataset Analysis.}
Previous work has primarily focused on understanding dataset characteristics. \citet{Carlini2019DistributionDT} has focused on outlier analysis in datasets, while 
\citet{Siddiqui2022MetadataAU} and \citet{Swayamdipta2020DatasetCM} focused on identifying different subsets within datasets using training dynamics. \citet{Nam2022SpreadSA} focused on improving worst-group accuracy of datasets. \citet{Luccioni2023StableBA} explored social biases in text-to-image systems. \citet{DeSilva2022TheVO} explored the value of OOD examples. \citet{Akiki2023SpaceriniPS} presents a qualitative analysis of large scale research datasets. \citet{Talukdar2021TrainingDB} has focused on identification and isolation of mislabelled data. Our work targeted a different element in large scale datasets---the importance of different modalities for each instance.

\newparagraph{Multimodal Dataset Analysis.} 
Previous work has primarily analyzed biases towards specific modalities in multimodal datasets by using existing models~\citep{winterbottom2020modality, Bitton2021DataEM, Hendricks2021DecouplingTR}. Specifically, \citet{winterbottom2020modality} trained partial-input models and used them to analyze the different modalities used in TVQA. Our approach, which relies on a seed human annotation, allows us to reach different and more reliable conclusions~(\cref{sec:dataset_analysis}).
Previous studies have employed human evaluation to discern biases towards specific modalities in their datasets~\citep{lei2018tvqa, agrawal2016vqa,tapaswi2016movieqa,Chao2017BeingNB,Castro2020LifeQAAR,Wang2021MultiSubsAL}. However, these investigations were limited to datasets containing only two modalities, and they did not expand their evaluation using automatic methods. 
Specifically, in the original TVQA paper, \citet{lei2018tvqa} provided partial inputs to workers. Our study focuses on three modalities, and goes further by examining the interaction of single-modality questions, making further observations about the data. 
\citet{Talmor2021MultiModalQACQ} and \citet{alamri2019audiovisual} created datasets with three modalities and used human evaluators for data assessment.
Our human analysis differs as we specifically analyze the required modality for each instance and consider the integration of multiple modalities, unlike their approaches which either treated modalities as input or focused solely on dialog quality without considering modality integration for individual questions. 
\citet{chen-etal-2020-hybridqa} created HybridQA, which contains two modalities, and relied on human assessments to analyze the answerability of questions across different modalities. As shown in \cref{sec:dataset_experimental_setup}, the reliability of human assessments can vary; they occasionally claim inability to answer based on a particular modality, when in fact, they can. By implementing an approach which includes testing human responses, we provide a more dependable evaluation of our dataset's modality distribution.


%% file: sections/6-limitations.tex
Our analysis is based on automatic tools trained on relatively small amount of data ($\sim$500 training instances, \cref{sec:dataset_experimental_setup}) to annotate 150K instances. The high costs of annotation prevents us from further expanding the initial annotation seed.
Nonetheless, our analysis shows that our classifiers are fairly accurate (74--81\%), that they generalize well to out-of-distribution, and finally, that more data doesn't necessarily improve performance (\cref{sec:dataset_experimental_setup}).

Our method is designed for classification-based tasks. Extending it to generation tasks is not straightforward, largely due to the challenges associated with evaluating human responses, which makes it hard to give binary solvable/insolvable labels to instances.

%% file: sections/7-conclusion.tex
We presented a two-step method for analyzing multimodal datasets, contributing to the ongoing discourse on data quality assessment. We proposed an approach that leverages a small seed of human annotation to identify important modalities in a dataset. We applied this approach to analyze the TVQA dataset and the \merlot model
. Our findings reveal that almost all TVQA questions can be answered using only one modality at a time. Moreover, there's no specific bias towards any modality in the dataset, differing from previous research. Additionally, we demonstrated that the \merlot model struggles with questions requiring the image modality but performs better with audio or text modality. We also highlighted the model's difficulty in speaker recognition. Finally, we collected 150 instances that require more than one modality to answer, and demonstrated that the model performs poorly on them. Our results enhance our understanding of dataset characteristics, as well as provide insights into the performance and limitations of AI models, highlighting the need for more robust multimodal modeling.

%% file: sections/8-Acknowledgements.tex
This work was supported in part by the Israel Science Foundation (grant no. 2045/21). We would like to extend our thanks to our colleagues from Roy Schwartz's lab at Huji for their valuable assistance with the pilot annotations. A special acknowledgment goes to Jeff Moskowitz for his continuous, thoughtful feedback, annotations, and support. We also express our appreciation to Eytan Siegel for his helpful feedback, as well as to Or Malka, Bar Madvil, Cheneil Clarke, and Adva Madvil for their annotations.

%% file: sections/Appendix.tex
\subsection{Dataset Supplementary Materials}
\label{app:dataset-supp}
\begin{enumerate}
\item Author statement: We bear all responsibility in case of violation of right in using our dataset. 
\item License: Dataset is licensed under CC-BY 4.0 license https://creativecommons.org/
licenses/by/4.0/legalcode.
\item The data and code are included in the supplementary material folder and will be released at a later date.
\item Intended uses: Our aspiration is for researchers to use our newly created TVQA test set and the workers' annotations in order to evaluate their multimodal models.
\item TVQA~\citet{lei2018tvqa} and \merlot~\citet{Zellers2022MERLOTRN} are both licensed under MIT License.

\end{enumerate}

\subsection{Human Annotation of TVQA}\label{app:workers-analysis}
\input{sections/appendix/1-human_annotation_TVQA}

\subsection{The Classifiers} 
\label{app:scorer-analysis}

\subsubsection{Ablations}
\input{sections/appendix/2a-Ablations}

\subsubsection{Training Size Analysis}
\input{sections/appendix/2b-small_data_size} \label{app:data-size-analysis}

\subsection{TVQA - Extra Analysis}
\input{sections/appendix/3-TVQA_extra_analysis}

\subsection{Creating Multi Modal Questions} \label{app:mmqa}
\input{sections/appendix/4-new_test_creation}

%% file: sections/appendix/1-human_annotation_TVQA.tex
\begin{figure}
    \centering   %
\setkeys{Gin}{width=\linewidth}
\begin{subfigure}{0.45\linewidth}
\includegraphics[width=\linewidth,scale=0.5,clip]{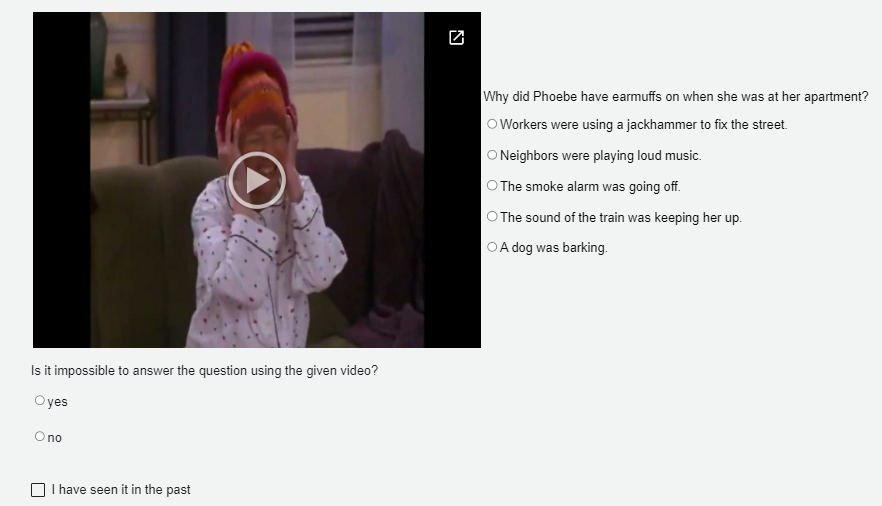} 
\caption{The Hit presented to the Image group}
\end{subfigure}%
\hfill
\begin{subfigure}
{0.45\linewidth}
\includegraphics[width=\linewidth,scale=0.5,clip]{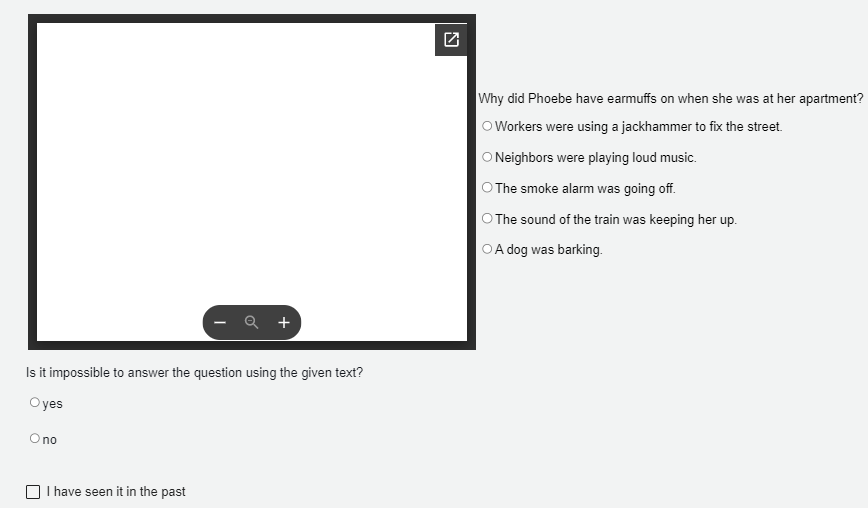} 
\caption{ The Hit presented to the Text group}
\end{subfigure}
\par\bigskip
\begin{subfigure}
{0.45\linewidth}
\includegraphics[width=\linewidth,scale=0.5,clip]{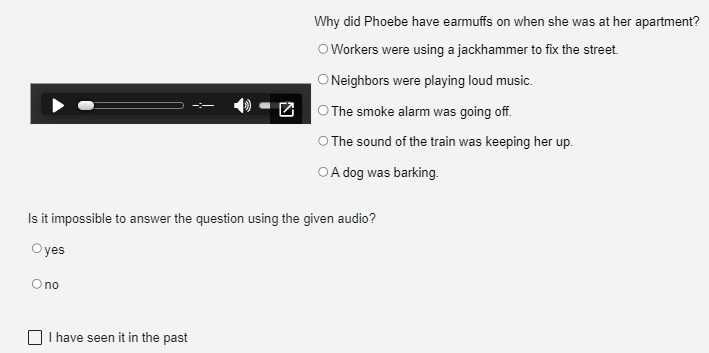} 
\caption{ The Hit presented to the Audio group}
\end{subfigure}
\caption{\label{fig:hit-example}
HIT examples from Mechanical Turk showcasing the annotation of different modality groups in TVQA.
}
\end{figure}
\begin{figure}
\includegraphics[width=\linewidth,scale=0.5, trim={4cm 4cm 4cm 4cm},clip]{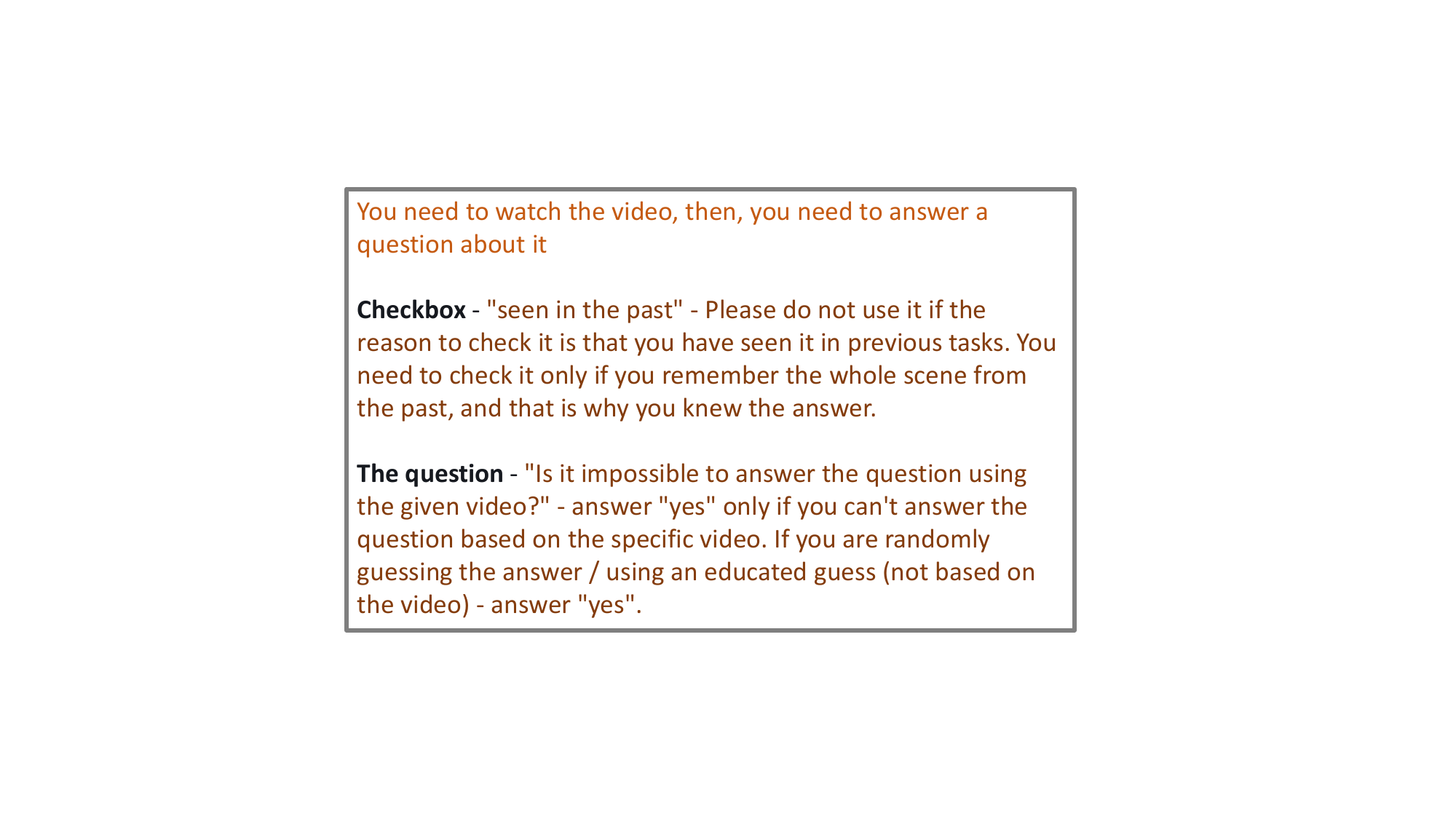} 
\caption{\label{fig:HIT-instruction} The Instructions presented to the image group when annotating the TVQA examples}
\end{figure}

Workers from various modality groups participate in HITs as depicted in Figure \ref{fig:hit-example}. They receive clear instructions on how to provide their answers. For instance, the image group is presented with instructions similar to the one shown in Figure \ref{fig:HIT-instruction}. The audio and text groups receive comparable instructions tailored to their respective modalities.

\newparagraph{Qualification.} 
The worker qualification process consists of two stages: an automatic stage followed by a manual stage. In the automatic stage, each modality group is assigned three HITs. Workers are required to answer questions based on the provided modality and indicate whether the question is answerable or not. Only workers who pass this initial stage proceed to the manual qualification process. In this stage, qualified workers annotate a batch of 10 HITs. Their annotations are evaluated based on agreement with other workers and ground truth. Workers who mark ``seen in the past" for more than 30\% of the questions are rejected. Additionally, the consistency between the signal of answering the question and the worker's response to ``Is it possible to answer this question based on the specific modality?'' is also assessed during evaluation.

\newparagraph{Payment.}
We hire nearly 20 workers, each worker is compensated at a rate of \$14-16 per hour for their participation in the project. The total expenditure for this project amounts to approximately \$1500.

\newparagraph{Selecting challenging instances.} \label{app:non-random} We select a set of 150 examples from the `Friends' section of the TVQA validation set, which exhibit the highest sensitivity to each modality in the \merlot model. This involved applying the model to the entire validation set and monitoring the probability of the gold label. We then run the model on inputs where each modality is masked. For each modality, we select the top 50 instances based on the largest decrease in the probability of the gold label when that modality is masked, compared to using all modalities as input. These instances may suggest questions that rely more heavily on a specific modality than others. Since there aren't many such instances for audio and text (due to the substantial overlap between these modalities in answering questions), this process makes the classifier training set more diverse for these modalities.

%% file: sections/appendix/2a-Ablations.tex
We experiment with various methods to recognize the data partitions that resulted in a decrease in performance when compared to human performance. These techniques highlight the importance of human annotations and model masking. \cref{tab:seventh,tab:fifth,tab:sixth} show the performance of the different approaches.
\begin{enumerate}
    \item  \textbf{Uni-modal models vs.~human annotators.} 
    
    We fine-tune the model on each subset of modalities five times using different seeds for each subset. During fine-tuning, all other modalities are masked. For instance, fine-tuning the model on the image modality involves masking audio and text in the video, allowing the model to receive only image frames as input for the question. For each subset and data point, we calculate the majority vote of the models to classify whether the example is solvable by that specific subset.
    
    \item  \textbf{Single model probabilities vs.~multiple masking modalities.} 
    
    We train a classifier using random forest, with various hyperparameters, which takes as input only the probability of the model given all modalities.
    
    \item \textbf{MLP vs.~random forest.} 
    
    We train a classifier using MLP with various parameters, using the same inputs as our previous classifier.
    
    \item \textbf{With gold label vs.~without gold label.} 
    
    We train a classifier similar to the one described previously, but without modifying the input to incorporate the gold label of the original question. 
\end{enumerate} 

\begin{table}[!htb]
\centering
\caption{\label{tab:seventh}Different approaches' performance for \textbf{audio} modality prediction, including OOD performance (150 examples from ``House'').}
\begin{tabular}{@{}ccccc@{}} \toprule
\textbf{Approach}    &  \textbf{Train}    &  \textbf{Val}    &  \textbf{Test}    &  \textbf{OOD}
\\ \midrule
majority & - & 69 & 69 & 71  \\
random & - & 50  & 50  & 50 \\ 
 1 - 5-models & 67  & 65  & \textbf{80} & \textbf{79}\\ 
2 - single probability & 62   & 71  & 67  & 71 \\
3 - mlp & 74   & 77  & 77  & 74 \\ 
4 - without gold label & 67   & 77  & 73  & 78 \\ 
our classifier & 84   & \textbf{81} & 76  & 77 \\ 
\bottomrule
\end{tabular}

\end{table}
\begin{table}[!htb]
\centering
\caption{\label{tab:fifth}Different approaches' performance for \textbf{text} modality prediction, including OOD performance (150 examples from ``House'').}
\begin{tabular}{@{}ccccc@{}} \toprule
\textbf{Approach}    &  \textbf{Train}    &  \textbf{Val}    &  \textbf{Test}    &  \textbf{OOD}
\\ \midrule
majority & - & 61 & 61 & 69 \\
random & - & 50  & 50  & 50 \\ 
 1 - 5-models & 70  & 66  & 65  & 77 \\ 
2 - single probability & 76   & 71  & 50  & 65 \\ 
3 - mlp & 78   & 78  &\textbf{ 77 } & 78 \\ 
4 - without gold label & 89   & 76  & 74  & \textbf{80 }\\ 
our classifier & 96   &\textbf{ 82 } & 74  & \textbf{80 }\\ 
\bottomrule
\end{tabular}

\end{table}
\begin{table}[!htb]
\centering
\caption{\label{tab:sixth}Different approaches' performance for \textbf{image} modality prediction, including OOD performance (150 examples from ``House'').}
\begin{tabular}{@{}ccccc@{}} \toprule
\textbf{Approach}    &  \textbf{Train}    &  \textbf{Val}    &  \textbf{Test}    &  \textbf{OOD}
\\ \midrule
majority & - & 72 & 72 & 61 \\
random & - & 50  & 50  & 50 \\
 1 - 5-models & 66  & 60  & 68  & 65 \\ 
2 - single probability & 97   & 75  & 72  & 61 \\ 
3 - mlp & 77   & 83  & 79  & 75 \\ 
4 - without gold label & 95   & 88  & \textbf{85 } & 79 \\ 
our classifier & 88   & \textbf{89 } & 81  & \textbf{80 }\\ 
\bottomrule
\end{tabular}

\end{table}

%% file: sections/appendix/2b-small_data_size.tex
\begin{table*}[tb]
\caption{Accuracy of training the classifiers with various amounts of data.}
\label{tab:three_tasks_results_different_training_size}
\centering
\begin{tabular}{@{}p{0.001cm}p{1.4cm}cccccccccc@{}} \toprule
\multicolumn{2}{l}{} & 
\multicolumn{3}{c}{Image} & 
\multicolumn{3}{c}{Text} & 
\multicolumn{3}{c}{Audio}  
\\ \cmidrule(l){3-5} \cmidrule(l){6-8} \cmidrule(l){9-11} 
\multicolumn{1}{l}{}  & Train & Val  & Test  & OOD   & Val  & Test  & OOD  & Val  & Test  & OOD \\ \midrule

& 30 \%              
& \phantom{0} 83              & \phantom{0} \textbf{85}                & \phantom{0} 83              & \phantom{0} 80              & \phantom{0} 71               & \phantom{0} \textbf{80}               
& \phantom{0} 78              & \phantom{0} \textbf{83}               & \phantom{0} 69  \\

& 50 \%
& \phantom{0} 85              & \phantom{0} 84                & 
\phantom{0} 81              & 
\phantom{0} 81              &
\phantom{0} \textbf{82}               & 
\phantom{0} 78               &
\phantom{0} 80              & 
\phantom{0} 79               & 
\phantom{0} \textbf{79}  \\

 & 70 \%
& \phantom{0} 84              & \phantom{0} 82                &
\phantom{0} \textbf{84}              & 
\phantom{0} \textbf{82}              & 
\phantom{0} 79       & 
\phantom{0} 79               & 
\phantom{0} 79              &
\phantom{0} 79               & \phantom{0} 75  \\

\midrule & 100 \%
& \phantom{0}\textbf{89}              & \phantom{0}81                &
\phantom{0}80              & 
\phantom{0} \textbf{82}              & 
\phantom{0} 74               & 
\phantom{0} \textbf{80}               & 
\phantom{0} \textbf{81}              &
\phantom{0} 76               & \phantom{0}77  \\
                            
\bottomrule
\end{tabular}

\end{table*}
 
The use of a small training size for the classifiers is beneficial as it simplifies the process of collecting data from workers, which can be both time-consuming and expensive. In this section, we show that even using a smaller training set size leads to roughly the same results. \cref{tab:three_tasks_results_different_training_size} displays the accuracy results for validation, test, and OOD when training the classifiers on  randomly selected subsets  of the training data. We conduct the same hyperparameter search on the validation set as done with our classifiers for each modality and amount of data. As shown in \cref{tab:three_tasks_results_different_training_size}, increasing the data size does not substantially improve performance, indicating that there is no added benefit in collecting more data.

%% file: sections/appendix/3-TVQA_extra_analysis.tex
\subsubsection{Analysis of TVQA based on workers}

To validate the findings from the analysis conducted on the entire validation set of TVQA, we perform similar experiments (as in \cref{sec:dataset_analysis}) on the annotated data  generated by workers. These experiments aim to assess the answerability of different modalities, evaluate the performance of the \merlot model on questions solvable by various modalities, and replicate the model's difficulty with image modality.

\begin{figure}[htbp]
\centering
\begin{subfigure}[t]{0.48\textwidth}
\includegraphics[width=\textwidth,scale=2, trim={1cm 0.3cm 1.1cm 0.5cm},clip]{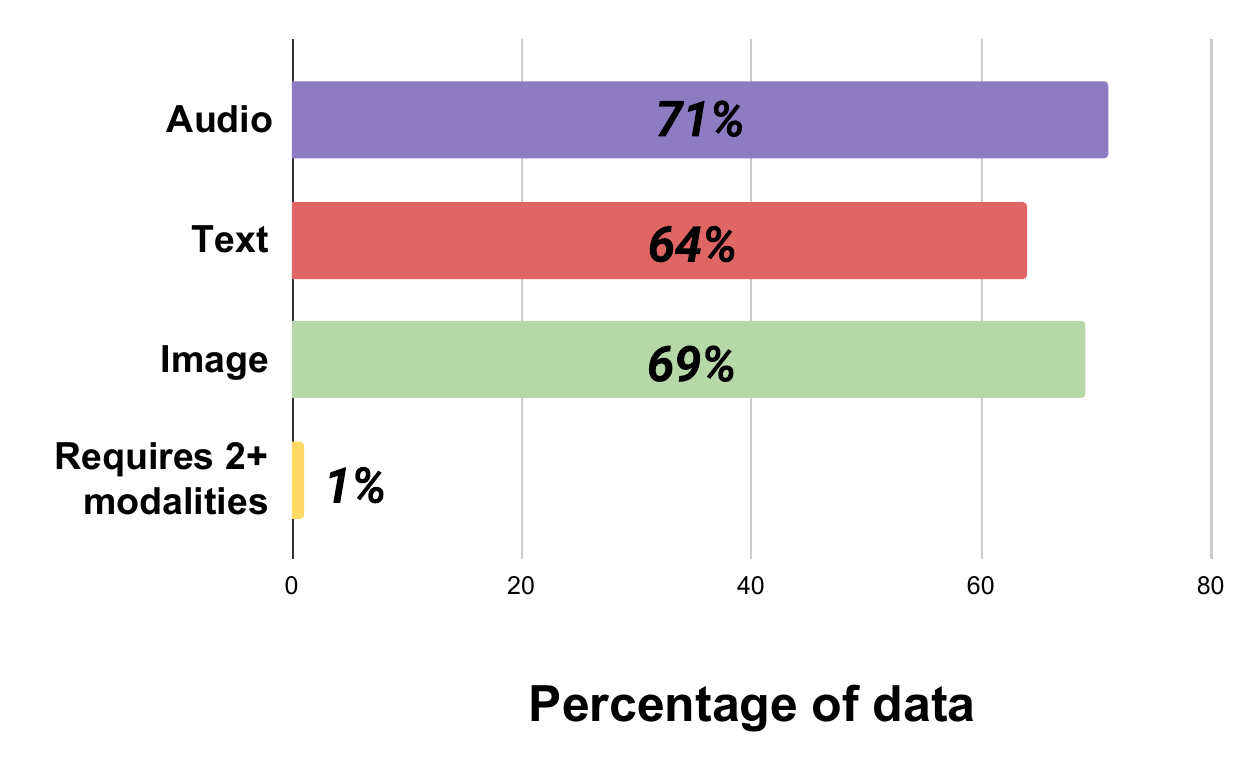}
    \caption{\label{Histogram} The proportion of TVQA questions that could be answered using a single modality, with each bar representing a different modality. The final bar represents the proportion of data that is unanswerable using any single modality.}
\end{subfigure}
\hfill
\begin{subfigure}[t]{0.48\textwidth}
    \includegraphics[width=\textwidth,scale=6, trim={4cm 2.5cm 7cm 1cm},clip]{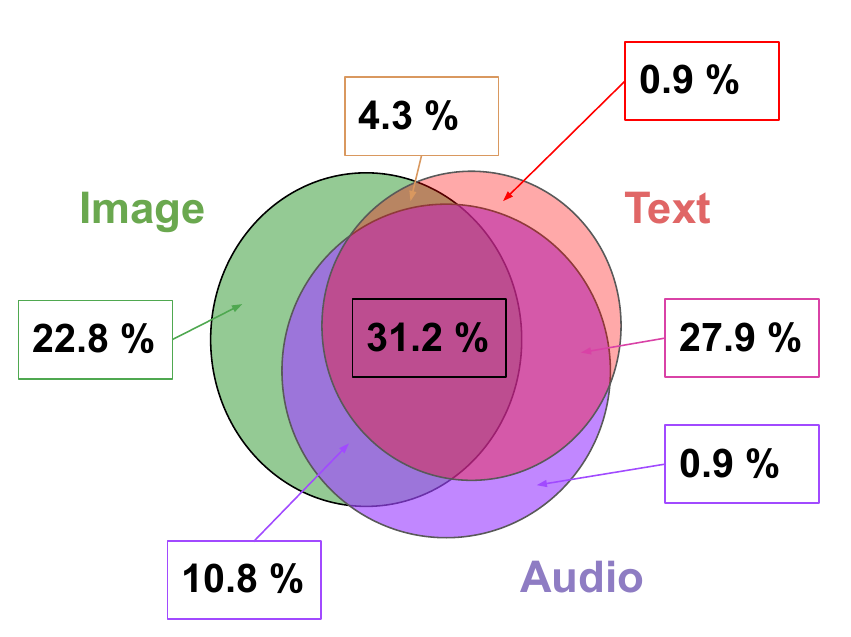}
        \caption{\label{all rand venn}
Each circle in the graph represents a portion of the data that can be solved only using a particular modality. Overlapping regions indicate the partitions that can be answered by either modality separately.}
\end{subfigure}
\caption{\label{fig:second}An analysis of our collected annotations of the TVQA dataset.}
\end{figure}

The data splits resulting from the classifiers applied to the annotated data by workers are shown in Figure \ref{fig:second}. These splits exhibit a similar trend to those observed in the validation data (Figure \ref{fig:validation-splits}), indicating that the annotations are representative. Furthermore, we apply the experiments described \cref{sec:dataset_model_analysis} to the annotated data. The results, presented in \cref{tab:seconda}, mostly replicate the findings from the full validation data. 

Additionally, the dataset cartography results obtained on the collected annotations set, as shown in the first block of Table \ref{tab:eight}, were consistent with those obtained from the annotated data. This further supports the notion that the analyzed human annotations are representative and applicable to the entire TVQA validation set.

\label{app:tvqa_analysis_based_on_workers}

\begin{table}
\caption{\label{tab:seconda} Accuracy scores of various partial-input models evaluated on different data splits derived from our collected annotations. Columns: Input modalities to the model, with others being masked (I:image, A:audio, T:text, I+A+T:all, A+T: audio and text). Rows: different splits evaluated in the study. Our analysis reveals a degradation in the accuracy of the image-only input model on various portions compared to other inputs or the full model. Moreover, the full model's performance on image-only questions is poor. We conclude that the model struggles with the image modality.}
\centering
\vspace*{0.5cm}
\begin{tabular}{@{}cccccc@{}} \toprule
\textbf{Data\textbackslash  Input}  & \textbf{I+A+T} &\textbf{A+T} 
& \textbf{I} &  \textbf{A} &  \textbf{T}
\\ \midrule 
All annotated data & 81 & -- &  52& 61 & 66 \\
Answerable-by-each  & 87 & -- & 57 & 68 & 79 \\
\midrule
Only Image & 74 & -- & 66 & -- & --\\ 
Only audio--text & 89 & 90 & --  & 86  & 87 \\
\bottomrule
\end{tabular}
\end{table}

\subsubsection{Extending the Analysis to the Training Data.} \label{app:halves_train}
\begin{figure}[t]
\centering
\begin{subfigure}[t]{0.46\textwidth}
\includegraphics[width=\textwidth,scale=2, trim={1cm 0.3cm 1.1cm 0.5cm},clip]{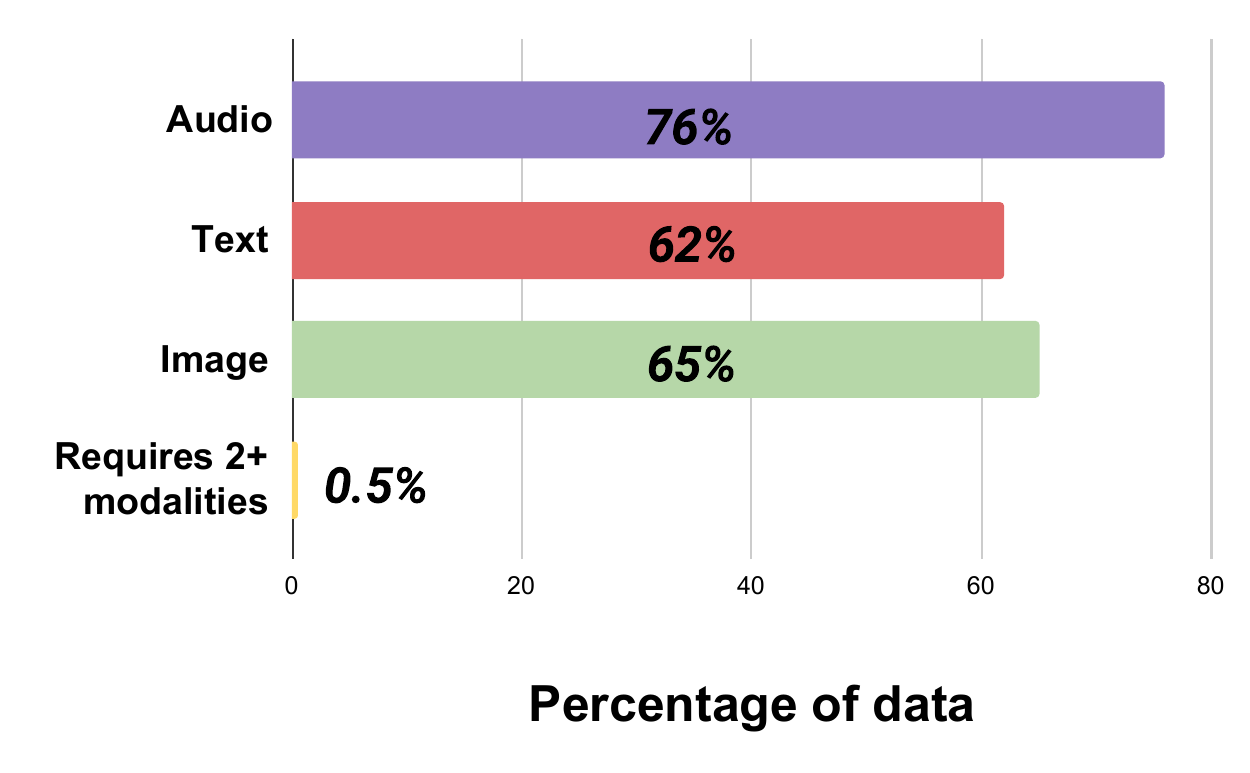}
    \caption{\label{train-hist}
    The graph displays the proportion of data that could be answered using different modalities such as audio, text, and image, with each bar representing a modality. The final bar represents the percentage of data that was unanswered by the workers for any modality.
    }
\end{subfigure}
\hfill
\begin{subfigure}[t]{0.46\textwidth}
\includegraphics[width=\textwidth,scale=6, trim={4cm 2.5cm 7cm 1cm},clip]{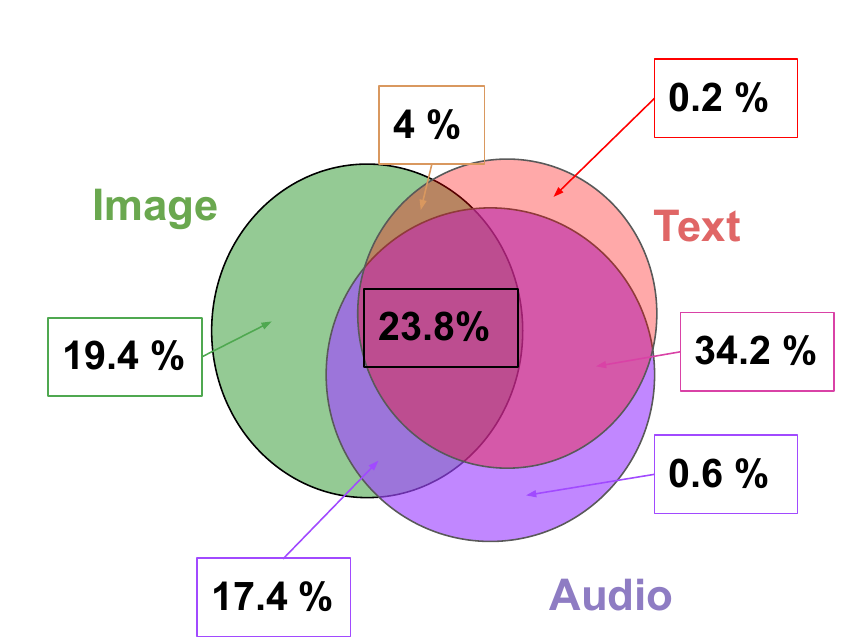}
        \caption{\label{train-venn}
Each circle in the graph represents a portion of the data that can be solved only using a particular modality. The overlapping region of the circles indicates the partitions that can be answered by either modality separately.
}
\end{subfigure}
\caption{\label{fig:splits}
An analysis of predictions of training split of the TVQA dataset.}
\end{figure}

After establishing that our predictions are reasonably accurate, we turn to explore the TVQA training dataset. 
Applying the same method here is not possible, since the training data is used to fine-tune our \merlot model. 
To address this, we randomly split the training data to two, and re-fine-tune \merlot twice, one for each half of the dataset. Subsequently, we train three classifiers with input representations from each of these models on the labels provided by humans. By doing so, we can extract probabilities from the model that is not trained on a specific half of the training set and apply the relevant classifiers to predict the modalities that can answer each instance in that half. This approach enables us to more accurately identify the data splits of the training set as if it were validation data. The resulting splits of the training data, as determined by the classifiers applied to the training set, are shown in \cref{train-hist,train-venn}. These splits are comparable to those observed in the workers' annotations analysis and to validation set, indicating a balanced distribution of the validation and training splits, which is valuable for generalizing the model. In an ideal scenario, it would be necessary to apply the classifiers to the test set of TVQA. However, since the gold labels for the test set are not available, and the classifiers relies on them, we are unable to perform this step.

\begin{table}[!htb]
\caption{\label{tab:eight}
Training dynamics analysis of our annotated portion of TVQA. For each data portion presented in rows, we calculate the proportion of questions answerable by each modality on it. First Block: Row (1): all annotated data portion, row(2): 50\% most ambiguous examples of the annotated data. Second Block: Row
(3): all training portion, row(4): 50\% most ambiguous examples of the training data.
The prevalence of image-based questions over other modalities in the 50\% most ambiguous questions indicates the model's difficulty with the image modality.}
\centering
\label{tab:table_swow_all_modalities}
\vspace*{0.5cm}
\begin{tabular}{@{}ccccccc@{}} \toprule
\textbf{Data}   &  \textbf{A} (\%)  &  \textbf{T} (\%)  & \textbf{I}  (\%) 
\\ \midrule
all annotated & 71 & 64 & 69 \\ 
most ambig.
 & 58 & 51 & 84\\
 \midrule 
all train & 76 & 62 & 65\\ 
most ambig.
& 64 & 42 & 90 \\ 
\bottomrule
\end{tabular}
\end{table}

%% file: sections/appendix/4-new_test_creation.tex
\subsubsection{Human Annotations}

\begin{figure}
\includegraphics[width=\linewidth,scale=0.5, trim={0cm 2cm 0cm 1.2cm},clip]{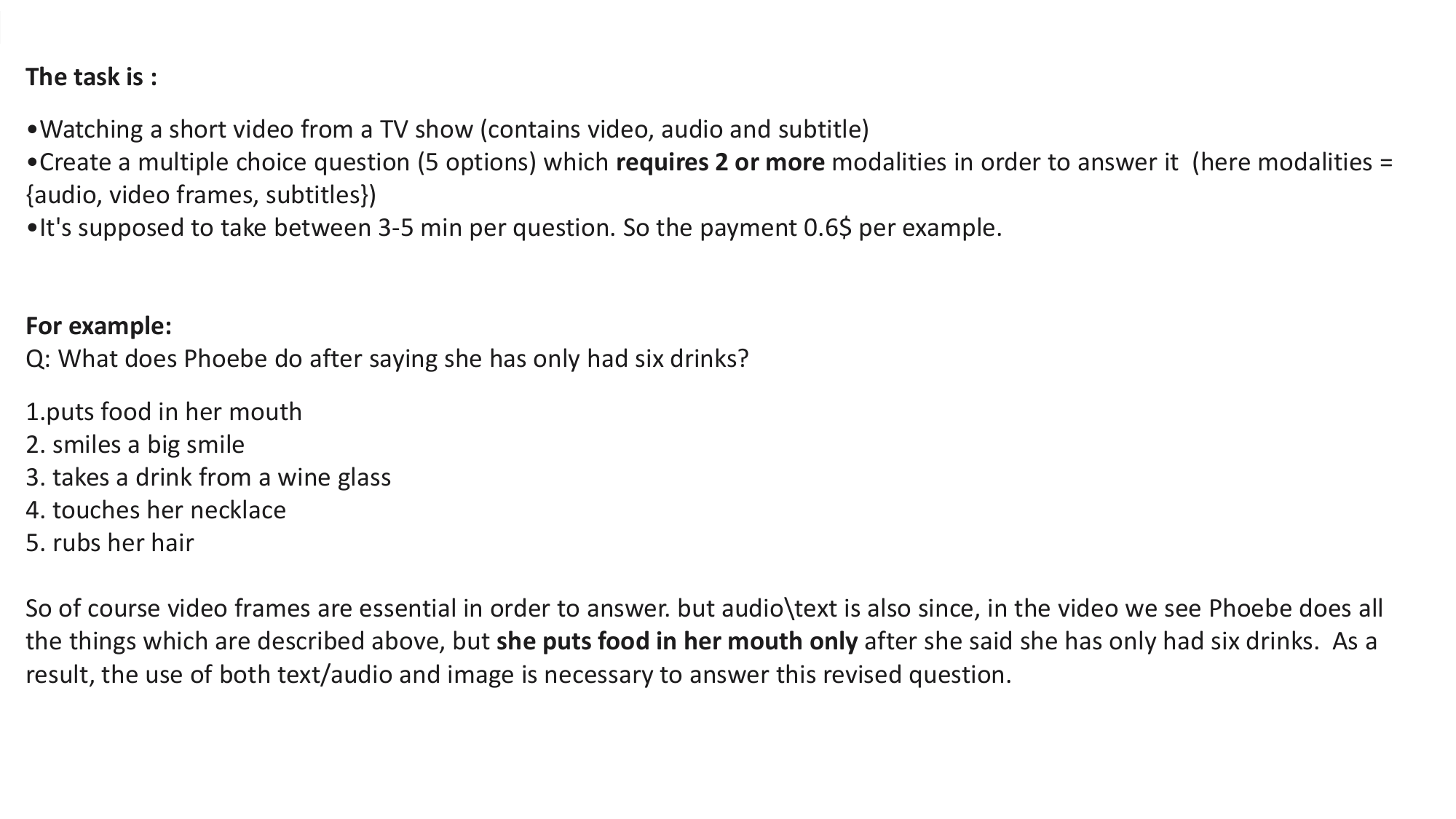} 
\caption{\label{fig:mmqa-instruction} The instructions for annotating the multi modal questions}
\end{figure}

We recruit workers from Amazon Mechanical Turk (AMT) who are proficient in English. These workers are provided with explicit instructions on how to create the questions, as depicted in~\cref{fig:mmqa-instruction}, to ensure clarity and consistency in their task.

\newparagraph{Qualification.}  Our crowdworkers undergo a qualification test that includes a small set of questions to confirm their ability to generate valid and answerable multimodal questions. As part of this test, each worker is provided with a set of existing videos from the TVQA dataset and instructed to create 10 questions that require the integration of at least two modalities for answering. Only those workers who successfully produce a minimum of 6 multimodal questions pass the qualification test.

\newparagraph{Data Collection.} Within the group of qualified workers, some are responsible for creating the multimodal questions, while others are assigned the task of validating them. Any questions that were not deemed multimodal or answerable were discarded during this validation process

\newparagraph{Payment.}
We hire 6 workers, each worker is compensated at a rate of \$14-16 per hour for their participation in the project.

\subsubsection{Extra Examples}
\begin{figure}[!htb]
\centering
\includegraphics[scale=0.7, trim={7cm 0cm 1cm 0.5cm},clip]{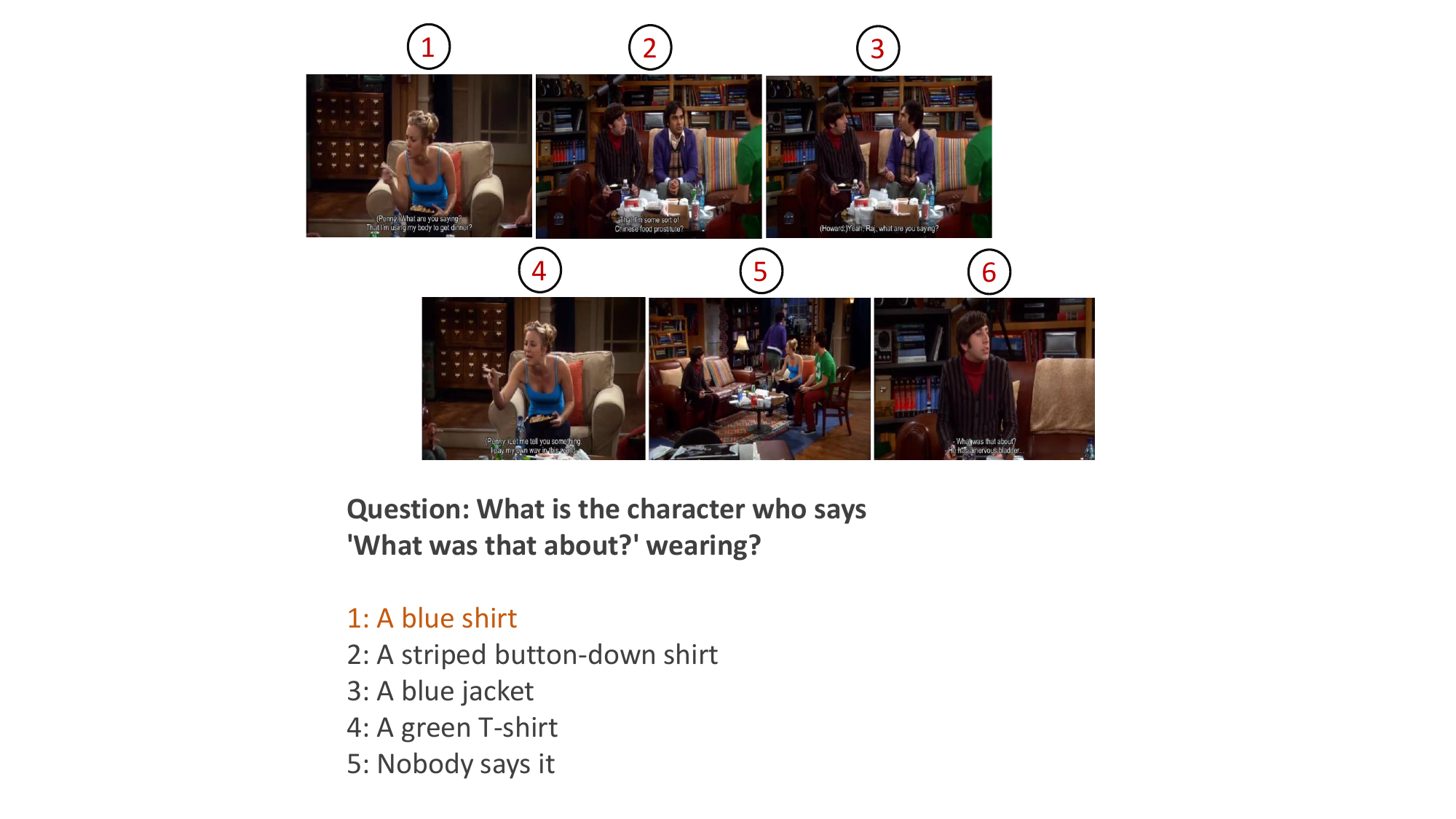}
\caption{\label{fig:mmqa2} Example no.1 of multi modal questions from our created test set. The video frames appear  in chronological order from left to right. This example is multimodal because it requires audio or text to determine that Penny said it, and then the image modality is needed to identify what she is wearing (other optional answers are clothing worn by others in the scene ).}
\end{figure}

\begin{figure}[!htb]
\includegraphics[scale=0.7, trim={7cm 0cm 1cm 0.5cm},clip]{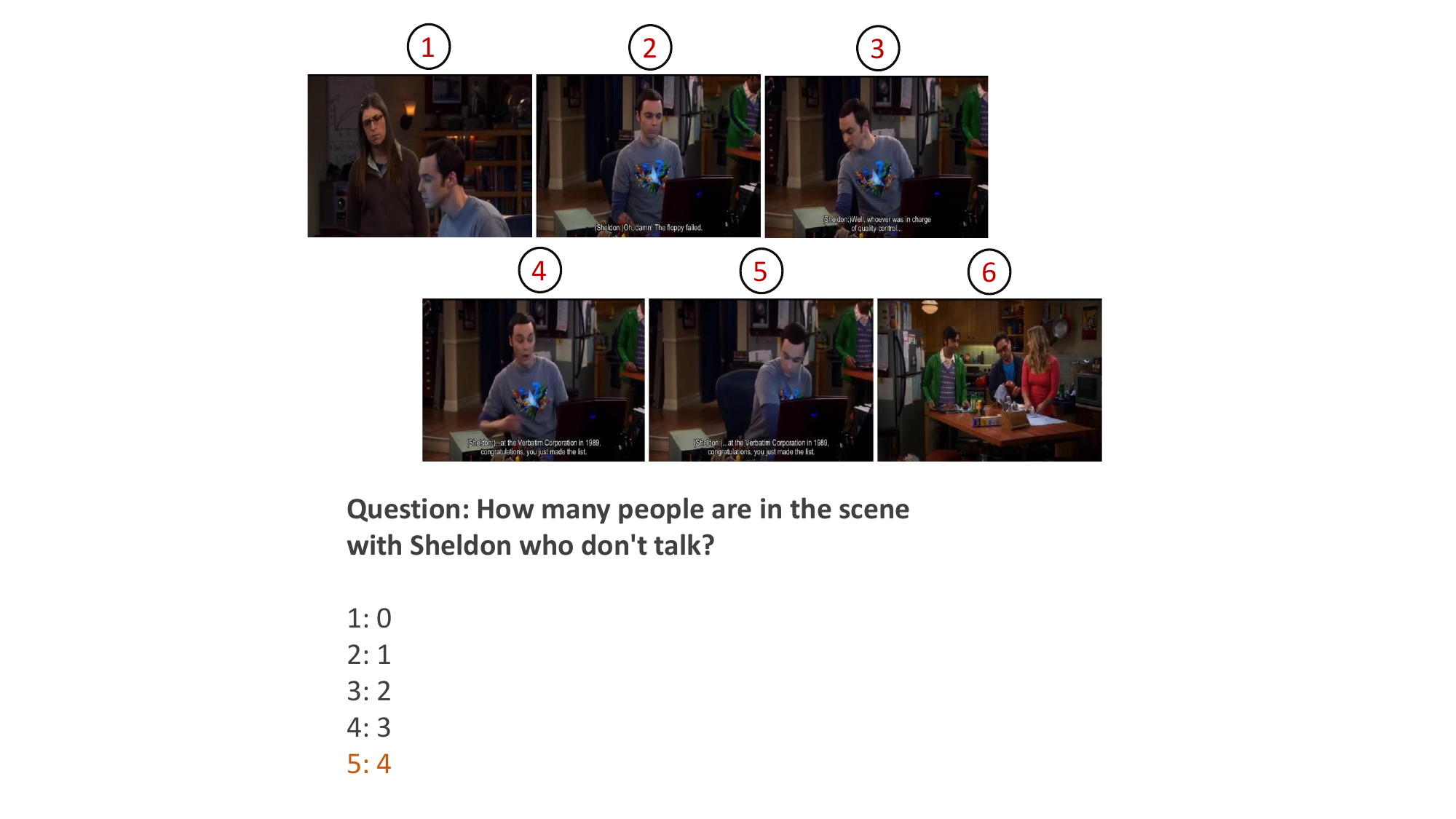}
\caption{\label{fig:mmqa3} Example no.2 of multi modal questions from our created test set. The video frames appear  in chronological order from left to right. This example is multimodal because it requires audio or text to determine how many people speak, and then the image modality is needed to see how many people are in the scene.}
\end{figure}

\begin{figure}[!htb]
\includegraphics[scale=0.7, trim={7cm 0cm 1cm 0.5cm},clip]{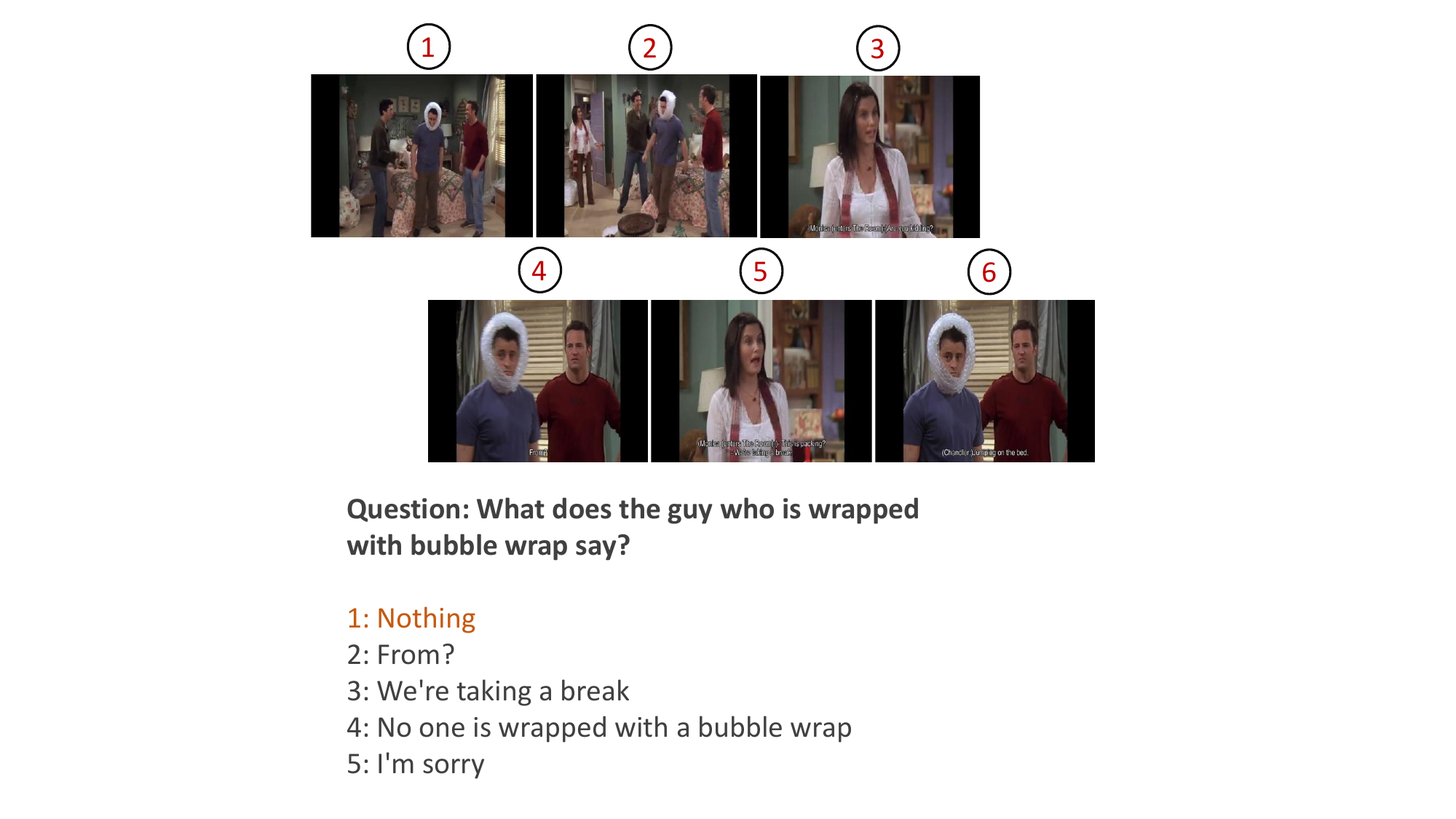}
\caption{\label{fig:mmqa4} Example no.3 of multi modal questions from our created test set. The video frames appear  in chronological order from left to right. This example is multimodal because it requires the image modality to identify Joey is wrapped with bubble wrap, and then audio or text are needed to determine he says nothing.}
\end{figure}
